\documentclass[journal,twoside,web]{ieeecolor2}
\usepackage{generic}
\usepackage{cite}
\usepackage{amsmath,amssymb,amsfonts}
\usepackage{algorithmic}
\usepackage{graphicx}
\usepackage{textcomp}
\def\BibTeX{{\rm B\kern-.05em{\sc i\kern-.025em b}\kern-.08em
    T\kern-.1667em\lower.7ex\hbox{E}\kern-.125emX}}
\usepackage{hyperref}
\hypersetup{hidelinks=true}
\usepackage{makecell}
\usepackage{stackengine}
\usepackage{multirow}
\usepackage{caption}
\usepackage{subfig}
\usepackage{colortbl}
\definecolor{gray}{rgb}{0.75,0.75,0.75}

\begin{document}
\title{Depth-Weighted Detection of Behaviours of Risk in People with Dementia using Cameras}

\author{Pratik K. Mishra, Irene Ballester, Andrea Iaboni, Bing Ye, Kristine Newman, Alex Mihailidis, and Shehroz S. Khan
\thanks{This paragraph of the first footnote will contain the date on which you submitted your paper for review. 
This work was supported by AGE-WELL NCE Inc, Alzheimer’s Association, Natural Sciences and Engineering Research Council, UAE Strategic Research Grant, Walter and Maria Schroeder Institute for Brain Innovation and Recovery and
the European Union’s Horizon 2020 research and innovation programme under MSCA grant agreement number 861091 for the visuAAL project.}
\thanks{Pratik K. Mishra, Alex Mihailidis and Shehroz S. Khan are with KITE, Toronto Rehabilitation Institute, Toronto, ON, Canada, and Institute of Biomedical Engineering, University of Toronto, Toronto, ON, Canada (e-mail: pratik.mishra@mail.utoronto.ca; alex.mihailidis@utoronto.ca; shehroz.khan@uhn.ca).}
\thanks{Irene Ballester is with Computer Vision Lab, TU Wien, 1040 Vienna, Austria (e-mail: irene.ballester@tuwien.ac.at)}
\thanks{Andrea Iaboni is with KITE, Toronto Rehabilitation Institute, Toronto, ON, Canada, and Department of Psychiatry, Temerty Faculty of Medicine, University of Toronto, Toronto, ON, Canada. (e-mail: andrea.iaboni@uhn.ca)}
\thanks{Bing Ye is with KITE, Toronto Rehabilitation Institute, Toronto, ON, Canada. (e-mail: bing.ye@utoronto.ca)}
\thanks{Kristine Newman is with Daphne Cockwell School of Nursing, Toronto Metropolitan University, Toronto, ON, Canada. (e-mail: kristine.newman@torontomu.ca)}}

\maketitle

\begin{abstract}
Background: The behavioural and psychological symptoms of dementia, such as agitation and aggression, present a significant health and safety risk in residential care settings. Many care facilities have video cameras in place for digital monitoring of public spaces, which can be leveraged to develop an automated behaviours of risk detection system that can alert the staff to enable timely intervention. However, one of the challenges is the presence of false alarms in these systems. Methods: We proposed a novel depth-weighted loss to enforce equivalent importance to the events happening both near and far from the cameras, aimed to reduce false alarms. We further propose to utilize the training outliers to determine the anomaly threshold. The data from nine dementia participants across three cameras in a specialized dementia unit were used for training. Ablation analysis was conducted for the individual components of the proposed approach and the effect of frame size and frame rate. The performance of the proposed approach was investigated for cross-camera, and participant-specific behaviours of risk detection. Results: The proposed approach obtained the best area under receiver operating characteristic curve performance of $0.852$, $0.81$ and $0.768$, respectively, for the three cameras. The proposed approach recorded a lower false positive rate than the existing methods, indicating its efficacy in reducing false alarms. Significance: This study motivates further research to make the behaviours of risk detection system more suitable for deployment in care facilities, improving the health and safety of people with dementia.
\end{abstract}


\section{Introduction} \label{sec_introduction}
%Dementia is a disorder characterized by growing impairment of cognitive functions such as memory, and reasoning and can affect insight, impulse control and judgement of a person \cite{henderson2000definition}.
%According to the World Health Organization 2023 report, there were 55 million people with dementia (PwD) worldwide, with nearly 10 million new cases every year \cite{who2023}. 
%The later stages of dementia are accompanied by the necessity of supervision and support for PwD in their everyday life, which is generally provided by long-term care (LTC) homes in the absence of support at home \cite{sloane2005evaluating}. 
% Approximately one-third of people living with dementia (PwD) under $80$ years in Canada live in long term care (LTC) homes, increasing to $42\%$ for those above $80$ years \cite{cihidementia}. 
People living with dementia (PwD) can develop behavioural and psychological symptoms putting themselves and the people around them at risk.
These behaviours of risk include pacing, pushing, hitting, kicking, intentional falling, and other behaviours that are self-harming, or harmful to others \cite{cohen1991instruction}. Long term care (LTC) homes often suffer from understaffing \cite{staffing} making it challenging for the staff to continuously oversee the safety of the residents. 
% Wearable sensors have been used in the past to detect clinically relevant episodes of agitation \cite{spasojevic2021pilot, iaboni2022wearable,cote2021evaluation}. However, they can be costly and intrusive, leading to dissent from PwD \cite{ye2019challenges}. These devices also need to be frequently removed and re-applied during bathing and battery charging.
%When an behaviours of risk detection system is deployed, depending on the number of residents in the LTC unit, it can be expensive to have a wearable device for each individual resident. 
Video cameras are an unobtrusive and inexpensive alternative for monitoring behaviour as they are already used by many LTC homes for security purposes, saving the time and expense of additional infrastructure installation.
Video cameras are rich in vital spatio-temporal information, which can be leveraged using deep learning approaches to design automated behaviours of risk detection systems that can be used to alert the staff in real-time and enable timely and appropriate interventions. This has important implications for safety and for helping to better understand the nature and triggers of the behaviours.
%Many care settings already have video cameras in place in common areas for security reasons, which can directly be integrated into the behaviours of risk detection system, saving the time and expense of additional infrastructure installation.

Behaviours of risk detection is a challenging problem due to their diversity and rare occurrence, making supervised classification approaches inappropriate.
An anomaly detection approach is more plausible, where a method can be trained to learn the normal behaviour characteristics and identify high variation cases as anomalies during testing \cite{mishra2021minimum}. In our previous work \cite{mishra2023privacy}, a spatio-temporal autoencoder was used to detect behaviours of risk as anomalies. However, there were significant issues with false positives (or false alarms), that could cause usability issues.
In general, video anomaly detection (VAD) autoencoder approaches can become sensitive to the background, leading to high reconstruction errors when a major part of the background is blocked \cite{liu2019exploring}. The objects or people closer to the camera cover a larger portion of the background than those further away, leading to significantly higher reconstruction errors and causing increased false positives. To tackle this problem, we propose a novel loss where the depth of the pixels is used as weight to diminish the overshadowing effect caused by the larger area of the people/objects close to camera. The aim is to reduce false positives by enforcing equivalent importance in analyzing the events happening in view irrespective of the distance to camera. In the absence of anomalies in the training set, we propose to utilize unusual activities that are not behaviours of risk, such as large objects or crowded scenes as proxy outliers to determine a threshold for detecting behaviours of risk. Expanding on our previous work \cite{mishra2023privacy} that analyzed behaviours of risk in one dementia participant using a single camera, here we extend the analysis to nine participants and three cameras.
The performance for detecting behaviours of risk can vary across individuals depending upon the type of behaviours of risk, and number of episodes recorded. To address this variability, we also conduct participant-specific analysis of the effectiveness of detecting behaviours of risk by the proposed approach. The main contributions of this work are:
\begin{enumerate}
    \item We propose an approach using a novel depth-weighted loss to enforce equivalent importance to events irrespective of distance from camera and utilize the training outliers to determine the anomaly threshold.
    \item We expand our analysis to nine participants with dementia and three different cameras.
    \item We integrate the proposed approach into three different existing anomaly detection methods, referred to as their depth variants. The effectiveness of the proposed approach is then evaluated by comparing the performance of the depth variants with the existing methods.
    \item We conduct an ablation analysis to investigate the effectiveness of individual components of the proposed approach and effect of frame size and frame rate in detecting behaviours of risk in PwD.
    \item We analyze the performance of the proposed approach for generalization across cameras, and participant-specific behaviours of risk detection in PwD.
\end{enumerate}

%The remainder of the paper is organized as follows: a brief literature review of existing work for behaviours of risk detection using videos is presented in Section \ref{sec_related}; the study details, data collection and demographic information of participants and proposed behaviours of risk detection method is presented in Section \ref{sec_method}. The results of experiments and related analysis are discussed in Section \ref{sec_results}; and, finally, conclusions and future work directions are provided in Section \ref{sec_conclusion}.

\section{Related Work} \label{sec_related}
In this section, we present the existing work for detecting behaviours of risk in PwD using videos and handling false positives in VAD scenarios.

\textbf{Behaviours of Risk Detection:}
The current approaches for detecting behaviours of risk in PwD use different types of modalities, including wearable devices \cite{spasojevic2021pilot}, computer vision \cite{fook2007automated,mishra2023privacy}, radio waves \cite{sharma2023wi} and multimodal / ambient sensing \cite{qiu2007multimodal}. As this work uses videos, further review only discusses the studies that either use videos alone or in combination with other sensors. 
% Fook et al~\cite{fook2006fusion} introduced a sensor fusion architecture which integrated ultrasound sensors, optical fiber grating pressure sensors, acoustic sensors, infrared sensors, radio-frequency identification, and video cameras. Bayesian networks were employed to model the uncertainties in sensor measurements. 
Qiu et al. \cite{qiu2007multimodal} proposed a multimodal information fusion model using various sensors, including pressure sensors, ultrasound sensors, infrared sensors, video cameras, and acoustic sensors. A layered classification architecture, consisting of a hierarchical hidden Markov model and a support vector machine, was employed. However, their results were based on mock-up data generated through simulation. 
Chikhaoui et al. \cite{chikhaoui2016ensemble} developed an ensemble learning classifier to detect agitation using data from a Kinect camera and an accelerometer.
% Ten participants were instructed to perform six different agitated and aggressive behaviours. 
However, the study did not specify whether the participants were healthy volunteers or PwD. 
Fook et al. \cite{fook2007automated} introduced a method employing a multi-layer architecture of a probabilistic classifier based on hidden markov model and a support vector machine classifier. However, the video data involved a person in bed, and it was unclear whether the participants were healthy individuals or PwD.
For all the above studies, it appears that these systems were conceptual and were never tested with real patients, so their performance is unknown in a real-world setting.
Acknowledging the importance of employing real-world data, we proposed an unsupervised convolutional autoencoder to detect agitation in PwD using videos collected from a specialized dementia unit from a single participant and a single camera \cite{khan2022unsupervised}. 
In our later work \cite{mishra2023privacy} with the same limited dataset, we presented privacy-protecting VAD approaches using body poses or semantic segmentation masks to detect behaviours of risk in PwD.

% \textbf{Video Anomaly Detection:}
% VAD is a task in computer vision, aimed at identifying unusual or anomalous events in video sequences in the absence of anomalous samples during training. Among these, autoencoders and depth-based methods have emerged as prominent approaches.
% % VAD using autoencoders
% Liu et al. \cite{liu2022appearance} proposed an appearance-motion united autoencoder framework for VAD which learned the appearance and motion normality jointly.
% Le and Kim \cite{le2023attention} proposed an attention-based residual autoencoder which encoded both spatial and temporal information in a unified way.
% % VAD using depth and Kinect
% Schneider et al. \cite{schneider2022unsupervised} used foreground-aware loss and investigated how existing algorithms could leverage depth modality from Kinect camera to enhance detection performance.
% Denkovski et al. \cite{denkovski2024temporal} proposed a multi-objective reconstruction and prediction-based loss to detect falls using multiple camera modalities including depth within a window of sequential frames.
% Yang et al. \cite{yang2023video} proposed a new task of video event restoration to infer missing frames motivating networks to mine and learn potential higher-level visual and temporal features for VAD.

\textbf{Reducing False Positives:}
A VAD method can land into usability issues in the presence of increased false positives. Therefore, it is important to assess the rate of false positives (or false alarms) in these applications. We briefly review existing methods that handle false positives for VAD scenarios. 
Doshi and Yilmaz \cite{doshi2021online} proposed an online VAD method and a procedure for selecting an operating threshold that satisfied a desired false alarm rate. The framework was composed of deep learning-based feature extraction from video frames and a statistical sequential anomaly detection algorithm.
Zhou et al. \cite{zhou2022vision} proposed an anomaly trajectory detection framework for online traffic incident alerts on freeways. It used an adversarial loss to enable the autoencoder to learn a better normal trajectory pattern that is beneficial for reducing false alerts. 
% , while an abnormal trajectory discriminator was established and trained to detect small mean shifts and filter out instantaneous false alerts.
Mozaffari et al. \cite{mozaffari2023self} proposed an online and multivariate anomaly detection method that was suitable for the timely and accurate detection of anomalies. They analyzed an asymptotic false alarm rate and provided a procedure for selecting a threshold to satisfy a desired false alarm rate.
Singh et al. \cite{singh2024cvad} 
% aimed to address the challenge of a high false alarm rate resulting from the trade-off between learning to reconstruct and distinguish anomalies. They 
proposed a constrained generative adversarial network where white gaussian noise was added to the input to boost the robustness of the model.

The existing methods tried to reduce false positives by leveraging algorithmic designs or choosing the threshold depending on a desired false positive rate. However, as described in Section \ref{sec_introduction}, the sensitivity to the distance of objects/people from the camera can be a determining factor in the increased false positives in VAD tasks, especially for scenes with large variations in depth, such as corridors or hallways. Hence, we utilize the depth of pixels to handle this issue and enforce equivalent importance to events happening far and close to the camera, thus reducing the false positives.

\section{Methods} \label{sec_dataset}
\subsection{Data Collection}
\begin{figure}[t]
\centering
\stackunder[2pt]{\includegraphics[width=0.32\columnwidth]{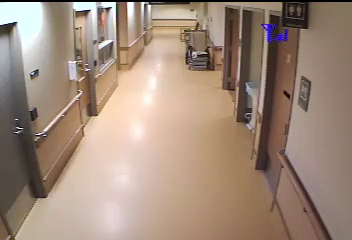}}{(a) Cam1}
\stackunder[2pt]{\includegraphics[width=0.32\columnwidth]{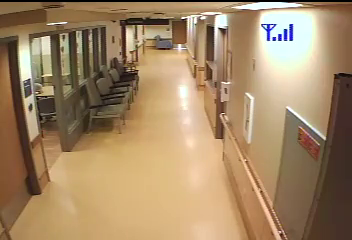}}{(b) Cam2}
\stackunder[2pt]{\includegraphics[width=0.32\columnwidth]{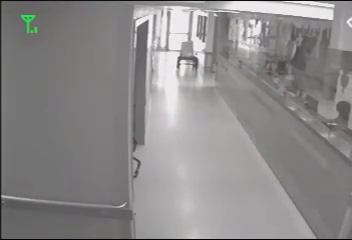}}{(c) Cam3}
\caption{Three different installed camera views.}
\label{fig_cams}
\end{figure}

The video data used in this work was collected as part of a larger multimodal sensor study \cite{spasojevic2021pilot,khan2024novel} at the Specialized Dementia Unit (SDU) at Toronto Rehabilitation Institute, University Health Network (UHN), Toronto, Canada, between November $2017$ and October $2019$. 
% The participants were admitted to the SDU from long-term care homes for the evaluation and management of behavioural and psychological symptoms in advanced dementia. 
Inclusion criteria encompassed individuals aged over 55 years, diagnosed with dementia and with a history of agitation or aggressive behaviours. 
% The participants who expressed dissent to use of a wearable device (one of the study sensors, not part of this video-based study) were excluded from the study. 
Data collection ceased under either of the following circumstances: absence of documented agitation for one week, sickness, death or discharge from the unit, or upon completion of two months of data collection. The data collection was approved by the research ethics board (UHN REB\#14-8483). Informed consent was obtained from substitute decision-makers for all participants. The staff also provided written consent for video recording within the unit. 
% Additionally, all staff and decision-makers of participants shown in the images in the paper have granted written consent for the publication of their images. 
For this study, we selected a subset of videos, approximately $93.5$ hours for each of the three cameras (see Fig. \ref{fig_cams}), based on 30 minutes before and after each documented behaviour of risk event. Table \ref{tab_demInfo} presents the demographic information of the participants. The selected video data captured the behaviours of risk events for nine dementia participants with an age range of $66$-$93$ years and a mean age of $81.22\pm8.12$ years. There were three male and six female participants.

\subsection{Data Annotation} \label{subsec_dataAnnt}
% It was observed that some of the behaviours of risk events in the selected videos were not documented previously as these events happened out of sight of the nursing staff. 
Two research assistants independently reviewed selected videos containing behaviour of risk events across Cam 1, 2 and 3, and annotated all the events corresponding to motor or physical agitation. 
% We had the timestamps of each behaviour of risk event, so the annotators did not have to watch the entire video feed but only look at the selected segment for each behaviour of risk event. This activity took approximately two months. 
Due to the lack of audio during data collection, it was not possible to annotate verbal aggression. Anvil \cite{anvil2012} tool was used for the annotation of camera videos. Both the research assistants also annotated the videos for other activities, namely, the presence of unusual activities, crowded scenes and large objects. The two assistants individually annotated $305$ and $383$ behaviours of risk events across three cameras and nine participants.
For the final labelling, an event was labelled as behaviour of risk if either of the research assistants annotated it as a behaviour of risk. For cases where both the assistants annotated an event as behaviour of risk but disagreed in the start/end times of the annotation, the duration with common agreement was labelled as a behaviour of risk. 
An inter-rater agreement analysis was performed, where Cohen's kappa \cite{mchugh2012}, Krippendorff alpha \cite{krippendorff1970} and percentage agreement metrics were calculated over the timestamps of the annotations of both research assistants. 
% This was done to measure the agreeability of the two annotations over the categorization of the events into normal and behaviours of risk events. 
The Cohen's kappa, Krippendorff alpha and percentage agreement were determined as $0.48$, $0.48$ and $0.99$, respectively. The value of Cohen's kappa and Krippendorff alpha can range from $-1$ to $1$, where $1$ and $-1$ represent a perfect agreement and disagreement, and $0$ represents agreement expected from random chance.

\subsection{Data Preprocessing}
The original resolution, frame rate and bit depth of the videos were $352\times240$, $30$ frames per second (fps) and $24$, respectively. The frames were sampled from the videos at $15$ fps, converted to grayscale, normalized to range $[0, 1]$ (pixel values divided by $255$) and resized to $64 \times 64$ resolution. This was done to minimize the computational expenses associated with trainable parameters. The frames were then stacked separately to form non-overlapping $5$-second windows ($75$ frames per window) (based on previous work \cite{khan2022unsupervised}). Training and test sets were generated individually for the three cameras. The training sets were composed of normal behaviours from daily activities only. The test set consisted of a combination of behaviours of risk events from nine participants and normal behaviours. 
% In the test sets, the available duration of annotated behaviours of risk events was limited. Hence, the duration of normal behaviour in test sets was chosen to keep a similar imbalance ratio (as in previous work \cite{mishra2023privacy}), which is approximately $7\%$ for behaviours of risk events. 
Table \ref{tab_setSize} shows the size of training and test sets for each of the three cameras. The test set was used to evaluate the performance of methods for detecting behaviours of risk.
The number of $5$-second windows and types of behaviours of risk across participants and cameras in the test set are presented in Table \ref{tab_agitPart}.

\begin{table}
\centering
\caption{Participants’ demographic information.}
\label{tab_demInfo}
\setlength{\tabcolsep}{3pt}
\begin{tabular}{|l|l|}
\hline
Attribute & Value \\ \hline
Number of Participants                                   & 9                                        \\
Median age of participants (years)                       & 82                                      \\
Mean age of participants (years)                         & 81.22                                      \\
Standard deviation of the age of participants (years)    & 8.12                                      \\
Range of age of participants (years)                     & 66–93                                     \\
Gender                                                   & \makecell[l]{Males (3)\\ Females (6)} \\
\hline
\end{tabular}

\bigskip

\caption{Size of training and test sets (in minutes).}
\label{tab_setSize}
\setlength{\tabcolsep}{3pt}
\begin{tabular}{|l|c|c|c|}
\hline
& Cam1 & Cam2 & Cam3 \\ \hline
Train set					& 1225.25	& 1242.33	& 1205.66 \\
Test set					& 209.16	& 291.92	& 96	\\
Normal behaviour (Test set)	& 194.08	& 270.5		& 89.83	\\
Behaviours of risk (Test set)		& 15.08		& 21.42		& 6.17	\\
\hline
\end{tabular}

\bigskip

\caption{Number of 5-second windows and types of behaviours of risk across participants and cameras in test set.}
\label{tab_agitPart}
\setlength{\tabcolsep}{3pt}
\begin{tabular}{|l|l|c|c|c|l|}
\hline
Participant-ID & Sex		& Cam1	& Cam2	& Cam3 	& Type of behaviour of risk \\ \hline
Participant1  & F		& 17	& 25	& 12	& \makecell[l]{Hitting, Pushing, Falling,\\Kicking, Grabbing, Throwing,\\Other} \\ \hline
Participant2  & F		& 3		& 3		& 1		& Pushing, Other \\ \hline
Participant3  & M		& 15	& 26	& 15	& \makecell[l]{Hitting, Pushing, Kicking,\\Grabbing, Other} \\ \hline
Participant4  & F		& 121	& 199	& 44	& \makecell[l]{Hitting, Pushing, Kicking,\\Grabbing, Throwing, Other} \\ \hline
Participant5  & F		& 0		& 0		& 1		& \makecell[l]{Hitting, Pushing, Kicking,\\Other} \\ \hline
Participant6  & M		& 17	& 0		& 1		& \makecell[l]{Hitting, Pushing, Kicking,\\Grabbing, Other} \\ \hline
Participant7  & F		& 8		& 0		& 0		& Pushing, Grabbing, Other \\ \hline
Participant8  & F		& 0		& 3		& 0		& Pushing, Other \\ \hline
Participant9  & M		& 0		& 1		& 0		& Hitting, Grabbing, Other \\ \hline
% \multicolumn{2}{|l|}{Total}		& 181	& 257	& 74	&  \\ \hline
\end{tabular}
\end{table}

\subsection{Behaviours of Risk Detection} \label{sec_borDet}
In our previous work \cite{khan2022unsupervised, mishra2023privacy}, we used a customized spatio-temporal convolutional autoencoder (3DCAE) to detect behaviours of risk from data from one participant and a single camera. 3DCAE learned to reconstruct the input videos with normal behaviour during training. Hence, during testing, normal instances were expected to have a lower reconstruction error, while instances with high reconstruction error were treated as anomalies (behaviours of risk in our case).
However, it showed high false positives due to sensitivity to activities such as crowded scenes, large objects, or other events happening near the camera due to reasons described in Section \ref{sec_introduction}. As such, there was a need to handle this issue to make the method suitable for real-world deployment. Hence in this work, we propose a novel depth-weighted loss, where the depth of the pixels is used as a weight factor to enforce equivalent importance to both behaviours of risk and non-behaviours of risk events happening far and near the camera. It is hypothesized that the weighted-depth loss will lead to a reduction in false positives. The proposed approach consists of the novel depth-weighted loss and utilizing the training outliers to determine the threshold.

\textbf{Depth-weighted Anomaly Score:}
%The scoring of an anomaly should remain consistent regardless of its proximity to the camera. 
Existing autoencoder methods~\cite{mishra2023privacy} employ reconstruction error to detect anomalous frames. This strategy inherently favours anomalies closer to the camera, as they occupy more pixels in the image plane. This bias poses a challenge, especially in scenes with highly varying distances from the camera, such as corridors (see Fig. \ref{fig_cams}). As a result, anomalies at different distances receive unequal anomaly scores, with those closer to the camera being unfairly favoured.
To address this problem, we propose a loss where depth per pixel is used as weight during the calculation of reconstruction error to enforce equivalent importance to both near and far pixels.

\begin{figure}[t]
\centering
\stackunder[2pt]{\includegraphics[height=2.7cm]{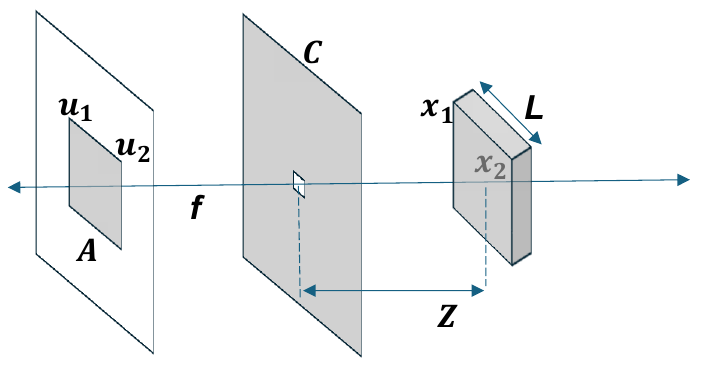}}{(a)}
\stackunder[2pt]{\includegraphics[height=2.7cm]{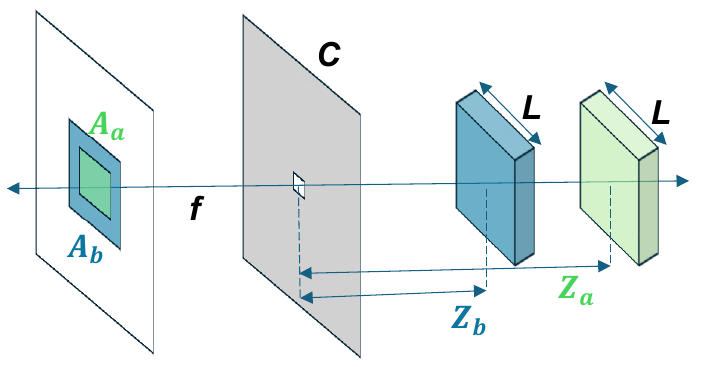}}{(b)}
\caption{(a) Illustration of the geometric relationship between a 3D prism (side length $L$) and the area of its projection in the image plane. (b) Two identical 3D prisms placed at different distances from camera position $C$ produce different projected areas. In this example, if $Z_a > Z_b$, then $A_a < A_b$.}
\label{fig_depth}
\end{figure}

To understand the translation of an anomaly in $3$D space, we describe the relationship between the area \( A \) depicting the anomaly in the image plane and its $3$D volume. For simplicity, we consider this $3$D volume as a square prism (see Fig. \ref{fig_depth} (a)). Consider two of the vertex of the square prism represented as \(x_1=(X_1, Y_1, Z_1)\) and \(x_2=(X_2, Y_2, Z_2)\) in the $3$D space and their $2$D projections \(u_1=(U_1, V_1)\) and \(u_2=(U_2, V_2)\) in the image plane. In Fig. \ref{fig_depth}, $C$ refers to the camera position. The projection of a $3$D point \((X, Y, Z)\) onto a $2$D image plane is expressed by the image coordinates \((U, V)\) through the pinhole camera model \cite{sturm2021pinhole} equations as,

\begin{equation}
U = \frac{fX}{Z} , \quad V = \frac{fY}{Z}
\label{eq_pinhole}
\end{equation}

where \( f \) is the focal length and \( Z \) is the distance from the $3$D object to the camera position. Assuming the difference in depth \( Z \) between these vertices is negligible (\( Z \approx Z_1 \approx Z_2 \)), the projections on the image plane according to eq. (\ref{eq_pinhole}) are:

\begin{equation}
U_1 = \frac{fX_1}{Z} , \quad U_2 = \frac{fX_2}{Z}
\end{equation}

Given that the length between the vertices in the 3D space along the X-axis is \( L = |X_2 - X_1| \), the area \( A \) of the square in the $2$D plane containing the anomaly is:

\begin{equation}
A = (U_2 - U_1)^2 = \left(\frac{f}{Z}\right)^2 (X_2 - X_1)^2 =\left( \frac{f L}{Z} \right)^2
\label{eq_area} 
\end{equation}

To analyze the sensitivity of the anomaly score $(AS)$ on the distance from the camera and therefore, on the area in the image plane, we can express the anomaly score of an area \( A \) as,

\begin{equation}
AS_{A} = \overline{AS}_{\text{pixel}} \cdot A
\label{eq_as} 
\end{equation}

where \(\overline{AS}_{\text{pixel}}\) represents the average anomaly score per pixel, a constant value that depends solely on the anomaly itself, irrespective of its position in 3D space or its projection onto the image plane.

Consider an anomaly event occurring at different distances \( Z_a \) and \( Z_b \) from the camera. As it is the same event with same volume, it will have same value of \(\overline{AS}_{\text{pixel}}\) and length $L$ at both distances. As illustrated in Fig. \ref{fig_depth} (b), when projected in the image plane, this results in different area values \( A_a \) and \( A_b \) in the image plane, with \( A_a < A_b \) if \( Z_a > Z_b \). Since both anomalies have the same value of \(\overline{AS}_{\text{pixel}}\), according to eq. (\ref{eq_as}), the difference in area will result in a different anomaly score. To ensure depth invariance, we need the anomaly to have the same score regardless of its distance from the camera. This is achieved by introducing a factor \( K \):

\begin{equation}
K_a \cdot AS_{A_a} = K_b \cdot AS_{A_b}
\label{eq_as_inv}
\end{equation}

Performing square root on both sides and combining the equations (\ref{eq_area}), (\ref{eq_as}) and (\ref{eq_as_inv}), we can express eq. (\ref{eq_as_inv}) as :

\begin{align}
\sqrt{K_a \overline{AS}_{\text{pixel}} \left( f L \right)^2 \frac{1}{Z_a^2}} = \sqrt{K_b \overline{AS}_{\text{pixel}} \left( f L \right)^2 \frac{1}{Z_b^2}} \nonumber \\
\implies K_a^{'} \sqrt{\overline{AS}_{\text{pixel}}} f L \frac{1}{Z_a} = K_b^{'} \sqrt{\overline{AS}_{\text{pixel}}} f L \frac{1}{Z_b}
\end{align}

Therefore, given the constant value of $\sqrt{\overline{AS}_{\text{pixel}}}$ and $f L$, to achieve depth invariance, the factors $K_a^{'}$ and $K_b^{'}$ must satisfy the following conditions:

\begin{equation} \label{eq_KZ}
K_a^{'} = Z_a, \quad K_b^{'} = Z_b
\end{equation}

Based on eq. \ref{eq_KZ}, we propose the following novel loss where depth $Z$ is used as weight during the calculation of reconstruction error as mean squared error to enable the methods to enforce equivalent importance to both near and far pixels.

\begin{equation}
    L_{dmse} = \frac{1}{N_e} \sum_{l=1}^{W} \sum_{i=1}^{S} \sum_{j=1}^{S} Z_{l,i,j} \: \: (I_{l,i,j} - O_{l,i,j})^2
    \label{eq_loss}
\end{equation}

where, $Z$ is the depth weight, $I$ is the input pixel, $O$ is the reconstructed pixel, $S$ is the spatial size, and $W$ represents the number of frames in a window. $N_e = W \times S \times S$ is the total number of pixels in a window.
% Following eq. (\ref{eq_Z2}), the value of depth-based weight hyperparameter $\phi$ is determined using $Z^2$ scaling by leveraging cross-validation method on the proxy validation set in a similar way as determining the threshold.
The weighted-depth loss helped the method to treat both the near and far behaviours of risk events with similar importance, which in turn helped to reduce false positives. The dense prediction transformer \cite{ranftl2021vision} was used to estimate the depth map during both training and testing. During testing, the depth-weighted loss was used as anomaly score for detecting behaviours of risk.

\begin{figure}[t]
    \centering
    \centerline{\includegraphics[width=0.95\columnwidth]{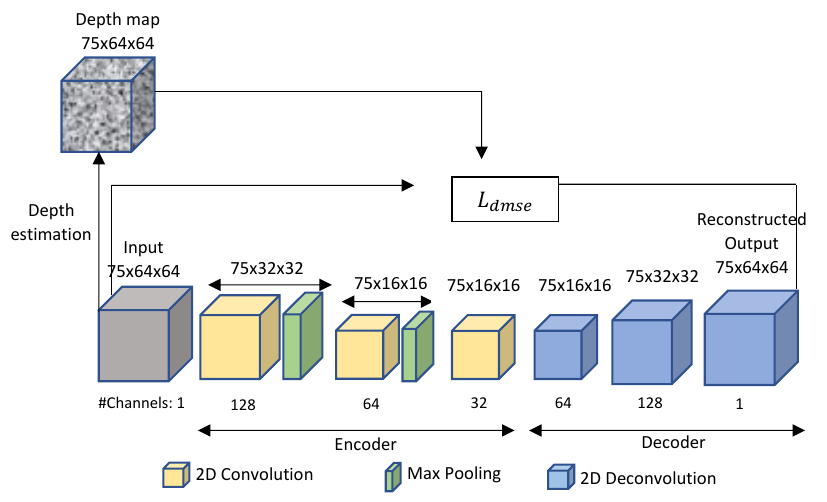}}
    \caption{depCAE architecture to detect behaviours of risk.}
    \label{fig_model}

    \bigskip
    
    \centering
    \includegraphics[scale=0.45]{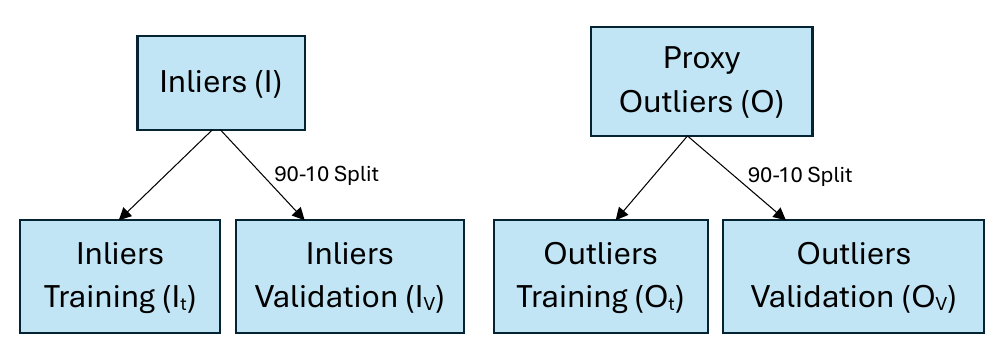}
    \caption{Proxy Validation Set Creation.}
    \label{fig_validationset}
\end{figure}

\textbf{Threshold Determination:}
In an anomaly detection approach, it can be challenging to determine an operating threshold due to the absence of anomalous samples in the training set. As discussed in Section \ref{subsec_dataAnnt}, the data was annotated for other unusual activities, such as large objects, crowded scenes and other infrequent activities in the training set. We propose to treat these activities as proxy outliers to create a proxy validation set to determine the threshold. We first create two sets: inliers ($I$) and proxy outliers ($O$). The set $I$ is composed of all the training windows. The set $O$ is comprised of windows annotated as other activities. Thereafter, $I$ and $O$ are further divided into (90\%-10\% split): training ($I_t$, $O_t$) and validation ($I_v$, $O_v$) sets \cite{khan2023empirical} (see Figure \ref{fig_validationset}). A method is trained on $I_t$, and the reconstruction error of all the windows in $I_t$ is determined. Then, each reconstruction error is chosen as an intermediate threshold, and the corresponding F1-score is calculated using the proxy validation set, which is the combination of $I_v$ and $O_v$. The reconstruction error with the highest F1-score on the proxy validation set is chosen as the threshold for testing.

\textbf{Experimental Setup:}
Transformers and variational autoencoders have performed well in the existing literature for VAD \cite{madan2023self, sun2018learning}. In this work, we have used three different anomaly detection methods, namely, 3DCAE, a self-supervised masked convolutional transformer block-based autoencoder (CTBAE) and a variational autoencoder (VAE) for evaluation. The 3DCAE method has been adapted from our previous work \cite{mishra2023privacy}. The CTBAE method used a self-supervised masked convolutional transformer block \cite{madan2023self} embedded after the encoder layer and its output was passed as input to the decoder. VAE \cite{PinheiroCinelli2021} learned to generate two vectors that represent the parameters (mean and variance) of a distribution from which the latent vector was sampled. For fairness during comparison, the encoder and decoder architecture of 3DCAE, CTBAE and VAE were kept the same and they used interquartile range (IQR) generated proxy-validation set \cite{khan2023empirical} for threshold calculation.
The proposed depth-weighted approach was applied as part of the 3DCAE, CTBAE and VAE methods and their depth variants are referred to as depCAE, depCTBAE and depVAE, respectively. The depCAE method followed an encoder-decoder architecture, where the encoder consisted of $2$D convolution and max-pooling layers, with a kernel size of ($1\times3\times3$), stride ($1\times1\times1$) and padding ($0\times1\times1$), followed by batch normalization and ReLU operations. The decoder used $2$D deconvolution layers ($2$D transposed convolution operation) with a kernel size of ($1\times3\times3$), strides ($1\times1\times1$), ($1\times2\times2$), ($1\times2\times2$) and paddings ($0\times1\times1$), ($0\times1\times1$), ($0\times1\times1$) for first, second, and third $2$D deconvolution layers, respectively. The overall architecture of the depCAE is presented in Fig. \ref{fig_model}. 
% In the depCAE method, $W$ = 75 and $N_e$ = $75\times64\times64 = 307200$.

\textbf{Evaluation Metrics:}
The proposed approach aims to reduce false positives; however, it is also important to consider false negatives when detecting behaviours of risk in PwD. Hence, we consider a metric, i.e. F1-score, that can give a balanced estimate of the performance on both false positive rate (FPR) and false negative rate (FNR). F1-score is the harmonic mean of precision and recall, and provides a reasonable performance estimate over true positive rate (TPR), FPR and FNR, simultaneously. Further, Matthew's Correlation Coefficient (MCC) gives an overall performance estimate over true positives (TP), true negatives (TN), false positives (FP) and false negatives (FN) and is expressed as, $MCC = (TP*TN - FP*FN)/\sqrt{(TP+FP) * P * N * (TN+FN)}$. Both F1-score and MCC are threshold-based metrics; hence they evaluate performance over a single threshold. The area under the curve of receiver operating characteristic (AUROC) evaluates the performance over TPR and FPR over a range of thresholds. Hence, the primary metrics employed for evaluating the performance of methods in this work are MCC, F1-score and AUROC. We also report additional metrics for reference, namely, TPR, true negative rate (TNR), FPR, and FNR. The performance is evaluated using depth-weighted loss as the score to identify behaviours of risk during testing.

\begin{table}[!htbp]
\centering
\caption{Performance comparison of depth variants and existing methods for Cam1.}
\label{tab_resCam1}
\setlength{\tabcolsep}{3pt}
\begin{tabular}{|l|c|c|c|c|c|c|}
\hline
\multirow{2}{*}{} & \multicolumn{3}{c|}{Existing methods}  & \multicolumn{3}{c|}{Depth variants} \\ \cline{2-7}
                  & 3DCAE & CTBAE & VAE   & depCAE         & depCTBAE       & depVAE         \\ \hline
TPR               & 0.945 & 0.950 & 0.641 & 0.939          & 0.79           & 0.541          \\
TNR               & 0.628 & 0.581 & 0.713 & 0.684          & 0.747          & 0.777          \\
FPR               & 0.372 & 0.419 & 0.287 & \textbf{0.316} & \textbf{0.253} & \textbf{0.223} \\
FNR               & 0.055 & 0.050 & 0.359 & 0.061          & 0.21           & 0.459          \\
% Gmean             & 0.770 & 0.743 & 0.676 & 0.801          & 0.768          & 0.649          \\
% Precision         & 0.165 & 0.150 & 0.148 & 0.187          & 0.195          & 0.159          \\
% Recall            & 0.945 & 0.950 & 0.641 & 0.939          & 0.79           & 0.541          \\
% Specificity       & 0.628 & 0.581 & 0.713 & 0.684          & 0.747          & 0.777          \\
% \makecell[l]{Balanced\\Accuracy} & 0.786 & 0.765 & 0.677 & 0.768          & 0.659          & 0.666          \\
MCC               & 0.301 & 0.276 & \textbf{0.198} & \textbf{0.335} & \textbf{0.305} & 0.191 \\
F1-score          & 0.281 & 0.259 & 0.240 & \textbf{0.313} & \textbf{0.313} & \textbf{0.246} \\
AUROC             & 0.844 & 0.779 & 0.754 & \textbf{0.852} & \textbf{0.821} & \textbf{0.784} \\ \hline
\end{tabular}
\bigskip
\caption{Performance comparison of depth variants and existing methods for Cam2.}
\label{tab_resCam2}
\setlength{\tabcolsep}{3pt}
\begin{tabular}{|l|c|c|c|c|c|c|}
\hline
\multirow{2}{*}{} & \multicolumn{3}{c|}{Existing methods}  & \multicolumn{3}{c|}{Depth variants} \\ \cline{2-7}
                  & 3DCAE          & CTBAE          & VAE   & depCAE         & depCTBAE & depVAE         \\ \hline
TPR               & 0.949          & 0.728          & 0.502 & 0.844          & 0.444    & 0.751          \\
TNR               & 0.611          & 0.668          & 0.812 & 0.686          & 0.819    & 0.719          \\
FPR               & 0.389          & 0.332          & \textbf{0.188} & \textbf{0.314} & \textbf{0.181} & 0.281 \\
FNR               & 0.051          & 0.272          & 0.498 & 0.156          & 0.556    & 0.249          \\
% Gmean             & 0.761          & 0.697          & 0.638 & 0.761          & 0.603    & 0.735          \\
% Precision         & 0.162          & 0.148          & 0.174 & 0.176          & 0.163    & 0.175          \\
% Recall            & 0.949          & 0.728          & 0.502 & 0.844          & 0.444    & 0.751          \\
% Specificity       & 0.611          & 0.668          & 0.812 & 0.686          & 0.819    & 0.719          \\
% \makecell[l]{Balanced\\Accuracy} & 0.780          & 0.698          & 0.657 & 0.632          & 0.622    & 0.735          \\
MCC               & \textbf{0.290} & \textbf{0.215} & 0.200 & \textbf{0.290} & 0.172    & \textbf{0.264} \\
F1-score          & 0.276          & \textbf{0.246} & 0.259 & \textbf{0.291} & 0.238    & \textbf{0.284} \\
AUROC             & \textbf{0.810} & 0.766 & 0.790 & \textbf{0.810} & \textbf{0.771}    & \textbf{0.792} \\ \hline
\end{tabular}
\bigskip
\caption{Performance comparison of depth variants and existing methods for Cam3.}
\label{tab_resCam3}
\setlength{\tabcolsep}{3pt}
\begin{tabular}{|l|c|c|c|c|c|c|}
\hline
\multirow{2}{*}{} & \multicolumn{3}{c|}{Existing methods}  & \multicolumn{3}{c|}{Depth variants} \\ \cline{2-7}
                  & 3DCAE & CTBAE & VAE            & depCAE         & depCTBAE       & depVAE \\ \hline
TPR               & 0.919 & 0.649 & 0.230          & 0.730          & 0.743          & 0.135  \\
TNR               & 0.530 & 0.645 & 0.903          & 0.714          & 0.622          & 0.931  \\
FPR               & 0.470 & \textbf{0.355} & 0.097 & \textbf{0.286} & 0.378          & \textbf{0.069}  \\
FNR               & 0.081 & 0.351 & 0.770          & 0.270          & 0.257          & 0.865  \\
% Gmean             & 0.698 & 0.647 & 0.455          & 0.722          & 0.680          & 0.355  \\
% Precision         & 0.118 & 0.111 & 0.139          & 0.149          & 0.119          & 0.119  \\
% Recall            & 0.919 & 0.649 & 0.230          & 0.730          & 0.743          & 0.135  \\
% Specificity       & 0.530 & 0.645 & 0.903          & 0.714          & 0.622          & 0.931  \\
% \makecell[l]{Balanced\\Accuracy} & 0.724 & 0.647 & 0.566          & 0.722          & 0.682          & 0.533  \\
MCC               & 0.220 & 0.149 & \textbf{0.105} & \textbf{0.235} & \textbf{0.182} & 0.063  \\
F1-score          & 0.210 & 0.190 & \textbf{0.173} & \textbf{0.248} & \textbf{0.205} & 0.127  \\
AUROC             & 0.765 & \textbf{0.739} & \textbf{0.709} & \textbf{0.768} & 0.722 & 0.696  \\ \hline
\end{tabular}
\end{table}

\section{Results and Discussion} \label{sec_results}
In this section, we evaluate the performance of the proposed approach for FP reduction and detecting behaviours of risk in PwD. We conduct an ablation analysis of different components of the proposed approach and the effect of frame size and frame rate on the performance of detecting behaviours of risk. We further perform cross-camera, and participant-specific analysis.

\textbf{Performance Evaluation:} We investigate the effectiveness of the proposed approach for reducing FP and detecting behaviours of risk by evaluating the improvement in performance after integrating the proposed approach into existing methods. As discussed in Section \ref{sec_borDet}, depth variants were created after integrating the proposed approach into the existing methods. Tables \ref{tab_resCam1}, \ref{tab_resCam2} and \ref{tab_resCam3} compare the performance of the depth variants with their existing method counterparts for behaviours of risk detection for Cam1, Cam2 and Cam3, respectively. The best values for FPR and primary evaluation metrics, MCC, F1-score and AUROC, between each method and its corresponding depth variant are marked in bold. It can be observed that depCAE performed mostly better than 3DCAE in terms of MCC, F1-score and AUROC for all three cameras. depCTBAE performed better than CTBAE for Cam1 and Cam3, while depVAE performed better than VAE for Cam1 and Cam2. depCAE recorded lower FPR than 3DCAE for all three cameras. depCTBAE and depVAE recorded lower FPR than CTBAE and VAE, respectively, for two out of three cameras. 
The methods performed better for Cam1 and Cam2 in comparison to Cam3, which can be attributed to the major portion of Cam3 view being occluded by the wall. This made it difficult to gauge the behaviours of risk events clearly in Cam3.
While depVAE gave lower FPR as compared to other methods, it gave significantly higher FNR. In such cases, it is important to consider a balanced evaluation over both FPR and FNR, which is provided by F1-score, where depCAE performed the best.
It can be inferred from the observations that the proposed approach of weighing the loss with pixel depth is indeed effective in reducing false positives and improving detection performance. Further, the proposed approach can be easily applied into any existing autoencoder-based anomaly detection method.
The depCAE method achieved a MCC, F1-score and AUROC of $0.335$, $0.313$, $0.852$ for Cam1, $0.29$, $0.291$, $0.81$ for Cam2, and $0.235$, $0.248$, $0.768$ for Cam3, respectively. The depCAE method achieved the best MCC, F1-score and AUROC overall for all three cameras; hence further analysis in this section is performed using depCAE method. 

% The behaviours of risk events occur rarely, leading to a skewed test set, where, only $7\%$ of test samples belong to the behaviours of risk class. As seen in Table \ref{tab_resCams}, the values of TPR and TNR are generally in the higher range and the values of FPR and FNR are generally in the lower range for depCAE method across three cameras. This suggests that given the imbalanced nature of the test set, the depCAE method performs well for detecting behaviours of risk in PwD. However, as the normal behaviour (or negative class) is present in a large majority in the test set, the effect of false positives is more prominent than true positives toward calculation of precision even if the TPR is high. This led to a relatively lower precision value, which in turn led to a lower F1-score value. 

\begin{table*}
\centering
\caption{Ablation analysis of proposed components of depCAE across three different cameras.}
\label{tab_resAblationComp}
\setlength{\tabcolsep}{3pt}
\begin{tabular}{|l|c|c|c|c|c|c|c|c|c|}
\hline
\multirow{2}{*}{Metric} & \multicolumn{3}{c|}{Cam1}					   & \multicolumn{3}{c|}{Cam2}							 & \multicolumn{3}{c|}{Cam3}						\\ \cline{2-10}
                  & depWgtOnly  & anntThrOnly  	& depCAE         & depWgtOnly  & anntThrOnly  		& depCAE 		& depWgtOnly  & anntThrOnly  & depCAE         \\ \hline
% Gmean             & 0.760       & 0.805      	& 0.801          & 0.743       & 0.791      		& 0.761         & 0.624       & 0.647      & 0.722          \\
% Precision         & 0.158       & 0.202      	& 0.187          & 0.158       & 0.188      		& 0.176         & 0.098       & 0.135      & 0.149          \\
% Recall            & 0.967       & 0.895      	& 0.939          & 0.879       & 0.907      		& 0.844         & 0.905       & 0.554      & 0.730          \\
% Specificity       & 0.598       & 0.725      	& 0.684          & 0.628       & 0.690      		& 0.686         & 0.430       & 0.756      & 0.714          \\
% Balanced Accuracy & 0.782       & 0.81       	& 0.811          & 0.754       & 0.798      		& 0.765         & 0.668       & 0.655      & 0.722          \\
MCC               & 0.294       & \textbf{0.344}      	& 0.335          & 0.269       & \textbf{0.325}      		& 0.290         & 0.167       & 0.172      & \textbf{0.235}          \\
F1-score            & 0.271       &\textbf{ 0.32} & 0.313          & 0.267       & \textbf{0.311}     & 0.291         & 0.177       & 0.217      & \textbf{0.248} \\
AUROC             & 0.851       & 0.842      	& \textbf{0.852} & 0.802       & \textbf{0.81}      & \textbf{0.81} & 0.765       & 0.768      & \textbf{0.768} \\ \hline
\end{tabular}
\end{table*}

\begin{table}[t]
\centering
\caption{Ablation analysis of input frame size and frame rates for depCAE.}
\label{tab_resAblationRestn}
\setlength{\tabcolsep}{3pt}
\begin{tabular}{|l|c|c|c|c|c|c|}
\hline
\multirow{2}{*}{Metric} & \multicolumn{2}{c|}{Cam1}		& \multicolumn{2}{c|}{Cam2}		& \multicolumn{2}{c|}{Cam3}	\\ \cline{2-7}
                  & s64r15  & s128r8  	& s64r15  & s128r8  & s64r15  & s128r8	\\ \hline
% Gmean             & 0.801 & 0.808 & 0.761 & 0.645 & 0.722 & 0.653 \\
% Precision         & 0.187 & 0.199 & 0.176 & 0.198 & 0.149 & 0.105 \\
% Recall            & 0.939 & 0.917 & 0.844 & 0.494 & 0.730 & 0.851 \\
% Specificity       & 0.684 & 0.712 & 0.686 & 0.841 & 0.714 & 0.501 \\
% Balanced Accuracy & 0.811 & 0.815 & 0.765 & 0.668 & 0.722 & 0.676 \\
MCC               & 0.335 & \textbf{0.345} & \textbf{0.290} & 0.226 & \textbf{0.235} & 0.173 \\
F1-score          & 0.313 & \textbf{0.326} & \textbf{0.291} & 0.282 & \textbf{0.248} & 0.187 \\
AUROC             & \textbf{0.852} & 0.848 & \textbf{0.810} & 0.795 & \textbf{0.768} & 0.739 \\ \hline
\end{tabular}

\bigskip

\caption{Cross-camera analysis for depCAE.}
\label{tab_resCross}
\setlength{\tabcolsep}{3pt}
\begin{tabular}{|l|c|c||c|c|}
\hline
                           & \makecell[c]{Train-test\\on Cam1} & \makecell[c]{Train on Cam2\\test on Cam1} & \makecell[c]{Train-test\\on Cam2} & \makecell[c]{Train on Cam1\\test on Cam2} \\ \hline
% Gmean             & 0.801                           & 0.750                                   & 0.761                           & 0.704                                   \\
% Precision         & 0.187                           & 0.162                                   & 0.176                           & 0.139                                   \\
% Recall            & 0.939                           & 0.862                                   & 0.844                           & 0.848                                   \\
% Specificity       & 0.684                           & 0.653                                   & 0.686                           & 0.585                                   \\
% \makecell[l]{Balanced\\Accuracy} & 0.811                           & 0.757                                   & 0.765                           & 0.717                                   \\
MCC               & 0.335                           & 0.274                                   & 0.290                           & 0.227                                   \\
F1-score            & 0.313                           & 0.272                                   & 0.291                           & 0.239                                   \\
AUROC          & 0.852                           & 0.796                                   & 0.810                           & 0.770  \\ \hline
\end{tabular}

\bigskip

\caption{depCAE performance stratified over participants.}
\label{tab_resPID}
\setlength{\tabcolsep}{3pt}
\begin{tabular}{|l|c|c|c|c|c|c|}
\hline
\multirow{2}{*}{} & \multicolumn{2}{c|}{Cam1}					   & \multicolumn{2}{c|}{Cam2}							 & \multicolumn{2}{c|}{Cam3}						\\ \cline{2-7}
             & F1-score  & AUROC  & F1-score  & AUROC  & F1-score  & AUROC  \\ \hline
Participant1 & 0.239   & 0.695  & 0.226   & 0.669  & 0.277   & 0.804  \\
Participant2 & 0.3     & 0.895  & 0.75    & 0.964  & - 	     & - 	  \\
Participant3 & 0.345   & 0.905  & 0.265   & 0.742  & 0.178   & 0.635  \\
Participant4 & 0.324   & 0.878  & 0.307   & 0.832  & 0.274   & 0.806  \\
Participant5 & - 	   & - 		& - 	  & -      & - 	     & - 	  \\
Participant6 & 0.276   & 0.784  & - 	  & -      & - 	     & - 	  \\
Participant7 & 0.32    & 0.819  & - 	  & -      & - 	     & -      \\
Participant8 & - 	   & - 		& 0.375   & 0.856  & - 	     & -      \\
Participant9 & - 	   & - 		& -       & - 	   & - 	     & -      \\ \hline
Average      & 0.3     & 0.829  & 0.385   & 0.813  & 0.243   & 0.748  \\ \hline
\end{tabular}

% \bigskip

% \caption{depCAE performance stratified over sex.}
% \label{tab_resSex}
% \setlength{\tabcolsep}{3pt}
% \begin{tabular}{|l|c|c|c|c|c|c|}
% \hline
% \multirow{2}{*}{} & \multicolumn{2}{c|}{Cam1}					   & \multicolumn{2}{c|}{Cam2}							 & \multicolumn{2}{c|}{Cam3}						\\ \cline{2-7}
% 			 & F1-score & AUROC  & F1-score  & AUROC  & F1-score  & AUROC  \\ \hline
% Male (N=2)    & 0.310  & 0.837  & 0.266   & 0.738  & 0.160   & 0.637  \\
% Female (N=4)  & 0.313  & 0.856  & 0.294   & 0.817  & 0.276   & 0.806  \\ \hline
% \end{tabular}
\end{table}

\textbf{Ablation Analysis I (Relative Effectiveness of the Proposed Components):} We investigate the relative effectiveness of the two components of the proposed method, namely, depth-weighted loss and threshold determination using annotated outliers, in detecting behaviours of risk in PwD. Table \ref{tab_resAblationComp} presents the results for ablation analysis over three cameras, where depWgtOnly refers to the setting of depCAE trained using depth-weighted loss while using IQR outliers \cite{khan2023empirical} for threshold calculation, and anntThrOnly refers to the setting of using mean squared error loss while using annotated outliers for threshold calculation. The bestMCC, F1 scores and AUROC are marked in bold. For Cam1, anntThrOnly performed better than depWgtOnly in terms of MCC and F1-score, while depWgtOnly performed better in terms of AUROC. Overall, depCAE benefited from the combined performance of depWgtOnly and anntThrOnly giving an AUROC of $0.852$. For Cam2, anntThrOnly performed similar to depCAE in terms of AUROC, while the performance of depCAE degraded in comparison to anntThrOnly in terms of MCC and F1-score. For Cam3, depCAE benefited from the combined functionality of depWgtOnly and anntThrOnly, outperforming both of them in terms of MCC, F1-score and AUROC. Overall, it can be observed that both depWgtOnly and anntThrOnly are vital components of depCAE and contribute significantly towards the performance of depCAE.

\textbf{Ablation Analysis II (Effect of Frame Size and Frame Rate)}: We investigate the effect of increasing the frame size and decreasing the frame rate in detecting behaviours of risk in PwD. Table \ref{tab_resAblationRestn} presents the results for ablation analysis over three cameras, where s64r15 refers to the original $64\times64$ frame size and $15$ fps setting, and s128r8 refers to $128\times128$ frame size and $8$ fps setting. For s128r8, a lower frame rate was used to manage the computation complexity added due to increase in frame size. The best MCC, F1 score and AUROC are marked in bold. It was observed that for Cam1, s128r8 performed better than s64r15 in terms of MCC and F1-score, while s64r15 performed better in terms of AUROC. For Cam2 and Cam3, s64r15 performed better in terms of all of MCC, F1-score and AUROC metrics. Overall, it was observed that s64r15 performed similar or better than s128r8. Increasing the frame size did not lead to any significant improvement in performance. The depCAE method can extract the necessary information for detecting behaviours of risk from a low resolution of $64\times64$, which may be difficult for the human eye.

\textbf{Cross-Camera Analysis}: We investigate the performance of depCAE to detect behaviours of risk in a new camera that was not seen during training. As part of this analysis, the depCAE method was trained on Cam2 and tested on Cam1 and vice-versa, as seen in Table \ref{tab_resCross}. The results are compared with the setting where the method is trained and tested on the same camera. It was observed that there was a drop in performance when the method was trained on a camera different from the testing camera. A drop in F1-score performance of $0.041$ ($0.313$-$0.272$) and $0.052$ ($0.291$-$0.239$), and in AUROC performance of $0.056$ ($0.852$-$0.796$) and $0.04$ ($0.81$-$0.77$) was observed when the method was tested on Cam1 and Cam2, respectively. The performance drop was expected and within reasonable limits as the training camera was different from the testing camera and no fine-tuning was involved. We have discussed some strategies to improve the cross-camera performance in Section \ref{sec_conclusion} as part of our future work.

\textbf{Participant-Specific Analysis:} As presented in Table \ref{tab_agitPart}, the type of behaviours of risk varied across participants. Hence, it is crucial to consider the potential dissimilar performance across individuals when training and testing behaviours of risk detection methods. Table \ref{tab_resPID} presents the performance of depCAE separately for each participant involved in the study across three cameras. ``-'' represents the cases where one or less than one agitation window was recorded for the participants. In general, it can be observed that depCAE showed better behaviours of risk detection performance for participants showing less diverse episodes of behaviours of risk, namely, participants $2$, $7$ and $8$, as compared to participants with more diverse episodes. depCAE performed generally well for participant $3$ across cameras $1$ and $2$ and across all three cameras for participant $4$, for whom the most number of behaviours of risk windows were recorded. On average, the detection performance was better for participants on cameras $1$ and $2$ as compared to camera $3$.

While the study presented important findings for the detection of behaviours of risk in PwD over multiple participants and cameras, there are some limitations associated with the study. First, as the SDU, where the data collection was conducted, usually admits PwD with severe cases of behavioural and psychological symptoms, the number of behaviours of risk episodes displayed by PwD can be higher than a standard LTC facility. Hence, it is crucial to revalidate these findings on the data collected in LTC facilities where the frequency of behaviours of risk episodes is expected to be lower. Second, due to privacy reasons, no camera was installed in private rooms, limiting the analysis of behaviours of risk episodes in public hallways. Hence, the applicability of our findings to more private settings remains unclear. Despite these limitations, our approach effectively detects behaviours of risk episodes with the best AUROC of 0.852, establishing it as the preferred option over existing methods.

\section{Conclusion and Future Work} \label{sec_conclusion}
We proposed an approach that uses a novel depth-weighted loss to enforce equivalent importance to both near and far events and uses training outliers to determine the anomaly threshold. The proposed approach, when applied to existing methods, recorded higher F1-score, MCC and AUROC while keeping lower FPR than the existing methods across all three cameras, thus performing better while reducing the false alarms in detecting behaviours of risk in PwD. The proposed approach consistently demonstrated its effectiveness and can easily be applied into any existing autoencoder-based anomaly detection method. This motivates further multidisciplinary research for the deployment of video surveillance-based behaviours of risk detection systems in nursing and care settings for PwD. Similar multidisciplinary approaches have shown to be helpful in medical settings \cite{manji2024effectiveness, basiri2021reduction}.

Future work can involve validating our approach on data collected from an LTC that typically encounters fewer behaviours of risk incidents. Further, the effect of utilizing the depth map as part of data as an additional input channel can be investigated.
Pose tracking cameras that directly extract body joints information without recording actual scene \cite{altumview} can be employed in private areas to detect behaviours of risk in a privacy-protecting manner \cite{mishra2023privacy,mishra2024skeletal}.
It can be a time-consuming and expensive task to train a method separately on each new camera scene or environment. Fine-tuning on target camera using few-shot learning and transfer learning can be employed to improve the performance of depCAE for cross-camera generalization.

In conclusion, this work marks an important step for the development of an automated behaviours of risk detection system that can be employed in real-world care settings and helps to improve the care and safety of PwD in assisted living facilities.

\bibliographystyle{IEEEtran}
\bibliography{ms}

\end{document}